\documentclass[11pt,a4paper]{article}
\pdfoutput=1 
\usepackage[hyperref]{acl2019}
\usepackage{times}
\usepackage{latexsym}
\usepackage{todonotes}
\usepackage{microtype}
\usepackage{xspace}

\usepackage{booktabs}
\usepackage{url}

\aclfinalcopy 
\def\aclpaperid{***} 

\newcommand{\CP}{\texttt{UCP}\xspace}

\title{
Pitfalls in the Evaluation of Sentence Embeddings}

\author{Steffen Eger\textsuperscript{$\dagger\ddagger$},~
Andreas R\"uckl\'e\textsuperscript{$\dagger$},~
\textbf{Iryna Gurevych\textsuperscript{$\dagger\ddagger$}}\\[.3em]
	\textsuperscript{$\dagger$}Ubiquitous Knowledge Processing Lab (UKP-TUDA)\\
	\textsuperscript{$\ddagger$}Research Training Group AIPHES\\
	Department of Computer Science, Technische Universit\"{a}t Darmstadt\\
	\textsuperscript{$\dagger$}{\url{www.ukp.tu-darmstadt.de}}\\
	\textsuperscript{$\ddagger$}{\url{www.aiphes.tu-darmstadt.de}}\\
}

\date{}

\begin{document}
\maketitle
\begin{abstract}
Deep learning models continuously break new records across different NLP tasks. At the same time, their success exposes weaknesses of model evaluation. Here, we compile several key pitfalls of evaluation of sentence embeddings, 
a currently very popular NLP paradigm. These pitfalls include the comparison of 
embeddings of different sizes, 
normalization of embeddings, and the low (and diverging) correlations between transfer and probing tasks. Our motivation is to challenge the current evaluation of 
sentence embeddings 
and to
provide an easy-to-access reference for future research. Based on our insights, we also recommend better practices for better  future evaluations of sentence embeddings. 
\end{abstract}

\section{Introduction}
The field of natural language processing (NLP) is currently in upheaval. A 
reason for this is 
the 
success story 
of deep learning, which has led to ever better reported performances across many different NLP tasks, sometimes 
exceeding the scores achieved by humans.
These
fanfares of victory 
are echoed by 
isolated voices raising concern about the 
trustworthiness of some of the  reported results. 
For instance, \citet{Melis:2017} find that neural language models have been misleadingly evaluated and that, under fair conditions, standard LSTMs outperform more recent innovations. \citet{Reimers:2017} find that reporting single performance scores is insufficient for comparing non-deterministic approaches such as neural networks. \citet{Post:2018} holds that neural MT systems are unfairly compared in the literature using different variants of the BLEU score metric.  
In an even more general context, \citet{Lipton:2018} detect several current ``troubling trends'' in machine learning scholarship, some of which refer to evaluation. 

%
%
Sentence encoders \cite{Kiros:2015,Conneau:2017,Pagliardini:2018} are one particularly hot deep learning topic. Generalizing the popular word-level representations \cite{Mikolov:2013,Pennington2014} to the sentence level, 
they 
are valuable in 
a variety of contexts: 
(i) 
clustering of sentences and short texts; (ii) 
retrieval tasks, e.g., retrieving answer passages for a question; and (iii) 
when
task-specific training data is scarce---i.e., when 
the full potential of task-specific word-level representation approaches cannot be leveraged \cite{Sandeep:2018}.

The popularity of sentence encoders has led to a large variety of proposed techniques. 
These 
range from `complex' unsupervised RNN models predicting context sentences \cite{Kiros:2015} to supervised RNN models predicting semantic relationships between sentence pairs \cite{Conneau:2017}. Even more complex models learn sentence embeddings in a multi-task 
setup \cite{Sandeep:2018}. 
In contrast, 
`simple' 
encoders
compute sentence embeddings as an elementary function of word embeddings. 
They
compute 
a weighted average of word embeddings 
and then modify these
representations
via principal component analysis (SIF) \cite{Arora:2017}; average n-gram embeddings (Sent2Vec) \cite{Pagliardini:2018}; 
consider generalized pooling mechanisms \cite{Shen:2018,Rueckle:2018}; 
or combine word embeddings via randomly initialized projection matrices \cite{Wieting:2019}. 

The embeddings of different encoders vary across various dimensions, the most obvious being their size. 
E.g., the literature has proposed embeddings 
ranging from 300d average word embeddings to 700d n-gram embeddings, to 4096d InferSent embeddings, to 24k dimensional random embeddings \cite{Wieting:2019}. 
Unsurprisingly, comparing embeddings of different sizes is unfair when size itself is crucially related to performances 
in downstream tasks, as has been highlighted before \cite{Rueckle:2018,Wieting:2019}. 

We compile several 
pitfalls when evaluating and comparing sentence encoders. These relate to (i) the embedding sizes, (ii) normalization of 
embeddings before feeding them to classifiers, and (iii) unsupervised semantic similarity evaluation. 
We also discuss 
(iv) the choice of classifier 
used on top of sentence embeddings 
and 
(v) 
divergence in performance results 
which compare 
downstream 
tasks and so-called 
probing tasks \cite{Conneau:2018}. 

Our 
motivation is to assemble diverse observations from different published works regarding problematic aspects of the emerging field of sentence encoders. We do so in order to provide future research with an easy-to-access reference about issues that may not (yet) be widely known. We also want to provide the newcomer to sentence encoders a guide for avoiding pitfalls that even experienced researchers have fallen prey to. We also recommend best practices, from our viewpoint. 



\section{Setup}
We compare
several freely available sentence 
encoders (listed in Table \ref{table:emb}) 
with SentEval \cite{Conneau:2018_lrec}, using its default settings. 
SentEval trains a logistic regression classifier for specific downstream tasks with the sentence embeddings as the input.
We compare 6 downstream tasks from the fields of sentiment analysis (MR, SST), product reviews (CR), subjectivity (SUBJ), opinion polarity (MPQA), and question-type classification (TREC). 
In these tasks, the goal is to label a single sentence with one of several classes. 
We also evaluate on the STSBenchmark 
\cite{Cer:2017}, which evaluates semantic similarity of \emph{pairs} of sentences.

\begin{table}[!htb]
    \centering
    \small
    \begin{tabular}{lr}
         \toprule
         \textbf{Sentence Encoder} & \textbf{Emb. Size} \\ \midrule
          InferSent \cite{Conneau:2017} & 4096 \\
          Sent2Vec  \cite{Pagliardini:2018} & 700 \\
          PMeans \cite{Rueckle:2018} & 3600 \\
          USE \cite{Cer:2018} & 512 \\
          Avg.~Glove \cite{Pennington2014} & 300 \\
          Avg.~Word2Vec \cite{Mikolov:2013} & ~~~variable \\
          SIF-Glove \cite{Arora:2017} & 300 \\
         \bottomrule
    \end{tabular}
    \caption{Sentence encoders used in this work, together with the sizes of the resulting sentence embeddings.}
    \label{table:emb}
\end{table}

\section{Problems}
\paragraph{Size matters.}
Currently, there is no standard size for sentence embeddings 
and different encoders induce embeddings of vastly different sizes.

For example, the sentence encoders of \citet{Conneau:2017}, \citet{Pagliardini:2018}, \citet{Rueckle:2018}, \citet{Cer:2018}, \citet{Kiros:2015}, \citet{Sandeep:2018} are 
4096, 700, 3600, 512, 4800, 1500/2048 
dimensional, respectively. 
However, \citet{Conneau:2017} show that their own model performs better when dimensionality of the embeddings is larger. 
They hypothesize
that the linear model they use for evaluation (logistic regression) performs better with higher dimensional embeddings because 
these 
are more likely to be linearly separable. \citet{Rueckle:2018} then argued that a comparison to low-dimensional baselines is unfair under this finding and increase the size of the baselines by concatenating different word embedding types or by concatenating different pooling operations (min, max, average). \citet{Wieting:2019} further extend this idea by enlarging the word embedding size with randomly initialized matrices before averaging. All three works show that performance increases as a concave function of embedding size, when a linear model is used on top of embeddings for evaluation.

\begin{figure}
\centering
\scalebox{0.85}{
\input{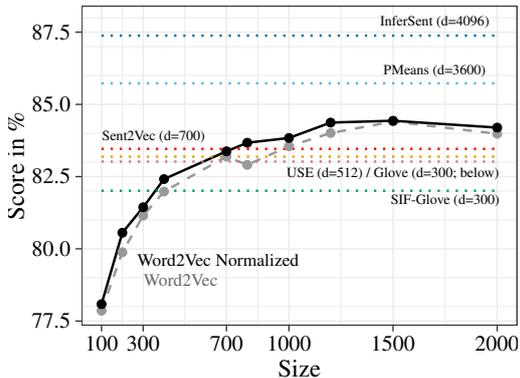}}
\caption{Avg. score across 6 transfer tasks for different sizes of Word2Vec embeddings vs.\ scores of other encoders (with constant embedding sizes as given in Table~\ref{table:emb}). 
`Word2Vec Normalized' is discussed below.
}
\label{fig:sizes}
\end{figure}

We also observe this trend when we merely train higher-dimensional word2vec word embeddings (on Wikipedia) and then average them, see Figure~\ref{fig:sizes}. 
At \emph{equal} embedding size, some models such as USE and Sent2Vec, have no or very little advantage over average word embeddings. 
Therefore, we strongly encourage future research to compare embeddings of the same sizes to provide a fair evaluation (or at least similar sizes).

\paragraph{Cosine similarity and Pearson correlation may give misleading results.}
The following evaluation scenario is common when testing for 
\emph{semantic similarity}:
given two inputs (words or sentences), embed each of them, (i) compute the cosine similarity of the pairs of vectors, and then (ii) calculate the Pearson (or Spearman) correlation with human judgments for the same pairs. Both 
steps are problematic: (i) 
it is unclear whether cosine similarity is better suited to measure semantic similarity than other similarity functions; 
(ii) Pearson correlation is known for its deficiencies---e.g., it only measures linear correlation and it is sensitive to outliers. A popular example for failure of Pearson correlation 
is Anscombe's quartet \cite{Anscombe:1973}.

Indeed, using such 
\emph{unsupervised} 
evaluations based on cosine similarity and Pearson correlation (which we denote \CP) may 
be misleading, as pointed out by \citet{Lu:2015}. 
When they normalized word embeddings,
their WS353 semantic similarity \cite{Finkelstein:2001} scores using \CP increased by almost 20 percentage points (pp). Since normalization 
is a simple operation that could 
easily be \emph{learned} by a machine learning model, this  indicates that \CP{} scores may yield unreliable conclusions regarding the quality of the underlying embeddings. 

We wanted to verify if this 
was also 
true 
when comparing different sentence encoders,
and therefore used the popular 
STSBenchmark dataset 
 in the \CP{} setup. 
The results in Table \ref{table:pearson} show that the outcomes vary strongly, with Glove-300d embeddings performing worst (0.41 correlation) and USE-512d embeddings best (0.70 correlation). We then normalized all sentence embeddings with z-norm\footnote{Subtracting the mean from each column of the embedding matrix of the whole data ($2N$ rows, one for each of $N$ pairs, and $d$ columns, where $d$ is the embedding size) and dividing by the standard deviation; after that we normalized each row to have unit length ($\ell_2$ norm).}, 
which is given as a recommendation in \citet{LeCun:1998} to
process inputs for deep learning systems. 

\begin{table}[!htb]
\centering
\footnotesize
\begin{tabular}{lcc|r}
\toprule
  & \textbf{Standard} & \textbf{Normalized} & $\Delta$ \\
  \midrule
  Glove-300d & 0.41 & 0.62 & +21 \\
  Word2Vec-300d & 0.56 & 0.65 & +9\\
  Word2Vec-800d & 0.56 & 0.67  & +11\\ 
  InferSent-4096d & 0.67 & 0.67 & +0\\
  SIF-Glove-300d & 0.66 & 0.67 & +1\\
  SIF-Word2Vec-300d & 0.67 & 0.67 & +0\\
  USE-512d & 0.70 & 0.70 & +0\\
  Sent2Vec-700d & 0.67 & 0.71 & +4\\
  PMean-3600d & 0.64 & 0.66 & +2\\
\bottomrule
\end{tabular}
\caption{Unsupervised cosine similarity + Pearson correlation (\CP) on STSBench (test data). 
$\Delta$ in pp. 
}
\label{table:pearson}
\end{table}

With normalization,
we observe a reduction in the range of the distribution of \CP{} scores across the nine systems 
from 29\% to 9\%; similarly, the standard deviation decreases from 8.6\% to 2.4\%. In particular, the worst sentence encoders catch up substantially: e.g., average Glove embeddings improve from 0.41 Pearson to 0.62 Pearson and the improvement of more complex systems over simple averaging baselines appears 
much less pronounced than before the normalization. 

When replacing Pearson by Spearman correlation, 
we observed very similar trends: 
e.g., Glove-300d had 0.44 before and 0.58 after normalization. 

This shows that
that 
\CP{} 
requires specific
normalization of the inputs for a fair comparison. 
Parallel to the suggestion of \citet{Lu:2015}, 
we recommend considering to use \emph{learned} similarity functions and mean-square error (MSE) 
as an alternative to 
\CP.

\paragraph{Normalization.}
Indeed, 
we 
also 
evaluated the effect of normalization
for \emph{supervised} transfer tasks, i.e., with learned similarity function. To this end, 
we compared the 9 sentence encoders from Table \ref{table:pearson} 
across 6 transfer tasks (averaged results) and STSBench (learned similarity function on training data instead of cosine similarity).\footnote{We estimated normalization vectors for mean and standard deviation on the training data and used these fixed values to normalize the test data.} 

\begin{figure}[!htb]
\input{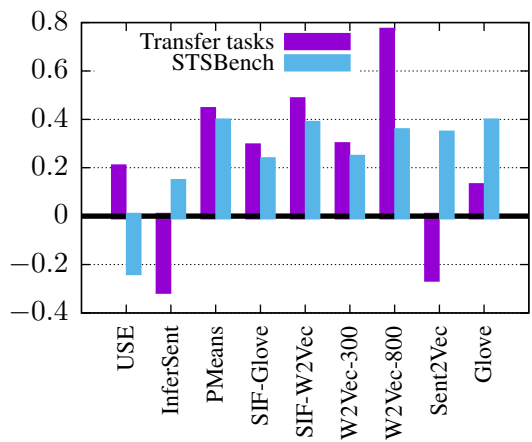}
\caption{$\Delta$ in pp of normalization and no normalization. Average across transfer tasks. STSBench differences are scaled down by factor of 10.}
\label{fig:normalization}
\end{figure}

Figure \ref{fig:normalization} shows that most techniques profit from normalization (on average), both on the transfer tasks and STSBench, 
even though gains are often substantially smaller than in the \CP{} setting.
Since the normalization can lead to rank changes (which we, e.g., observe between W2Vec-800, USE, and Sent2Vec) we thus  recommend introducing normalization as a binary hyperparameter that is tuned for each sentence encoder and downstream task also in supervised settings with learned similarity function.

\paragraph{Different classifiers for evaluation.} The popular SentEval evaluation tool feeds sentence embeddings into a logistic regression classifier. The underlying assumption is 
that in order to evaluate the embeddings \emph{themselves} the classifier used on top of embeddings should be as simple as possible. While the argument has some appeal, one wonders how relevant such an evaluation is when in practice more powerful classifiers would probably be used, e.g., deeper networks  
(current versions of SentEval also offer evaluating with a multi-layer perceptron (MLP)). {In particular, we note here an  
asymmetry between the extrinsic evaluation of word and sentence embeddings: word embeddings have traditionally been compared extrinsically by feeding them into different powerful architectures, such as BiLSTMs, while sentence embeddings are compared using the simplest possible architecture, logistic regression. While this is cheaper and focuses more on the embeddings themselves, it is less practically relevant, as discussed, and may have undesirable side effects, such as the preference for embeddings of larger size.}

A main problem arises when the ranking of systems is not stable across different classifiers. To our knowledge, this is an open issue. We are only aware of \citet{Sandeep:2018}, who evaluate a few setups both with logistic regression and using an MLP, and their results indicate that their own approach profits much more from the MLP than the InferSent embeddings they compare to ($+$3.4pp vs.\ $+$2.2pp). 

Thus, it is not sufficient to only report results with logistic regression, and evaluations with better-performing approaches would provide a more realistic comparison for actual use-case scenarios. 
We suggest reporting results for at least logistic regression and MLP. 

\begin{figure}[!htb]
\centering
\includegraphics[width=0.9\columnwidth]{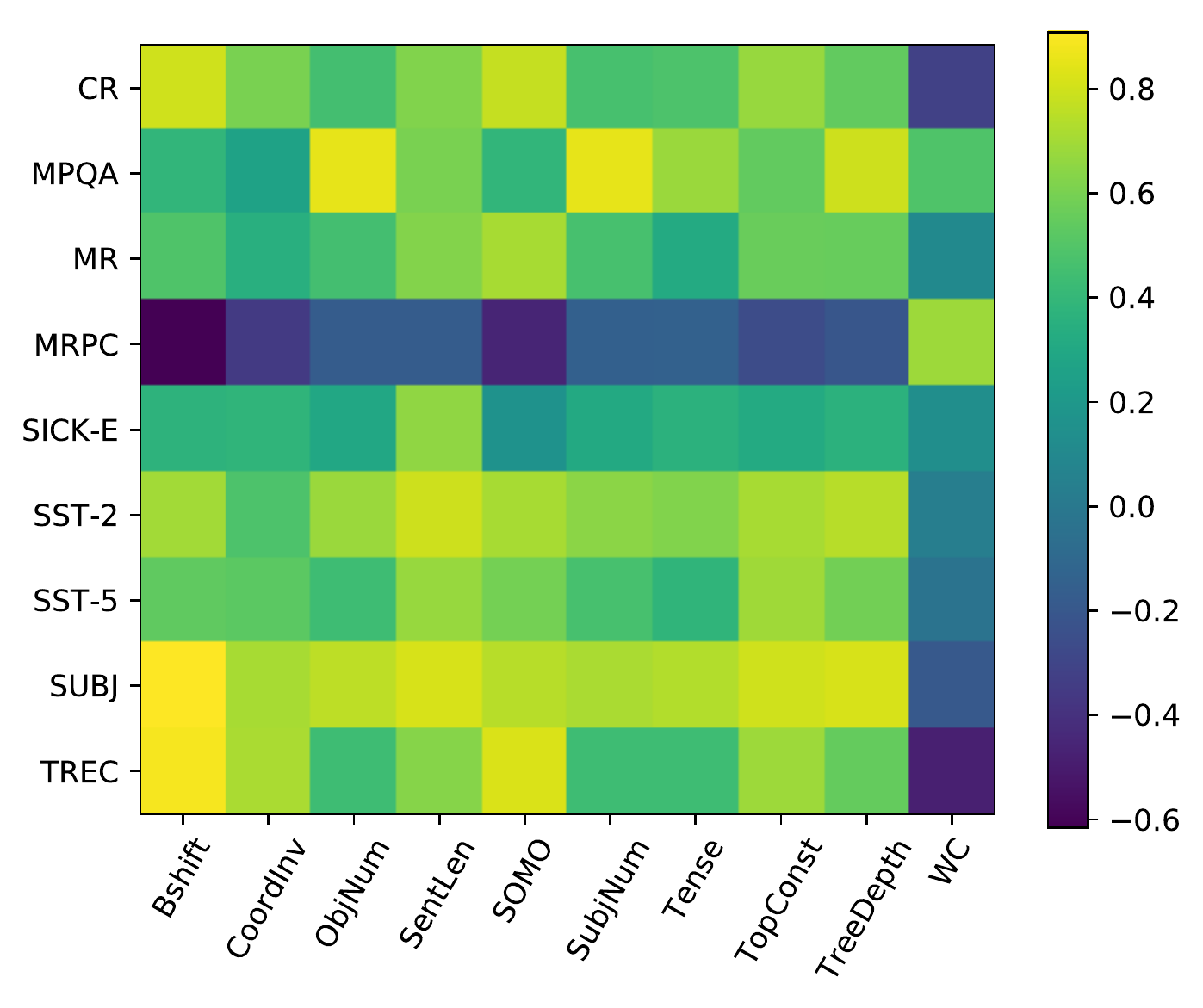}
\caption{Correlation of 11 sentence encoders on transfer tasks (y-axis) and probing tasks (x-axis) in \citet{Perone:2018}.}
\label{fig:probingtasks}
\end{figure}

\paragraph{Correlation of transfer tasks and probing tasks.} Besides transfer tasks, the literature has recently suggested evaluating sentence encoders with 
probing tasks \cite{Conneau:2018} that query embeddings for certain linguistic features, such as to detect whether a sentence contains certain words (WC) or to determine the sentence length (SentLen). 
\citet{Perone:2018} evaluate 11 different sentence encoders on 9 transfer tasks and 10 probing tasks. 
We plot the Spearman correlation between their transfer task results and their probing task results in Figure \ref{fig:probingtasks}. 
The average Spearman correlation is 0.64. The highest average correlation to transfer tasks has SentLen (0.83), and the lowest score has WC (0.04). Taken at face value, this may mean that current transfer tasks query more for superficial sentence features (knowing that embedding A can better predict sentence length than embedding B is indicative that A outperforms B on the transfer tasks) than for actual semantic content, as the embeddings were originally designed for. 

Thus, future research might focus on more suitable (difficult) datasets and sentence classification tasks for the evaluation of sentence embeddings, a lesson already learned in other fields \cite{Laubli:2018,Yu:2018}. 

Importantly, depending on the set of evaluated sentence encoders, such correlations can 
yield 
contradictory outcomes. For example, \citet{Conneau:2018} evaluate more than 40 combinations of \emph{similar}  sentence encoder architectures and observe the strongest correlation with downstream task performances for WC (cf. their figure 2). This is in contrast to the correlations from the results of \citet{Perone:2018}, where WC had the \emph{lowest} correlation.
Thus, it 
remains 
unclear to which extent downstram tasks benefit from the different properties that are defined by many probing tasks.

\section{Conclusion}



Others have laid out problems with the evaluation of \emph{word} embeddings  \cite{Faruqui:2016} using word similarity tasks. 
They 
referred to the vagueness of the data 
 underlying the tasks (as well as its annotations), 
the low correlations between extrinsic and intrinsic evaluations, and the lack of statistical tests. 
Our critique differs 
(in part) 
from this in that 
we also address 
extrinsic evaluation and 
the evaluation techniques themselves,\footnote{\citet{Faruqui:2016} also discuss some short-comings of cosine similarity.} 
and in that we believe that the comparison between sentence embeddings is not always fair, especially given the current evaluations using logistic regression. 
This
implicitly favors larger embeddings, and may therefore result in misleading conclusions regarding the superiority of different encoders. 
As practical recommendations, 
we
encourage future research in sentence embeddings to (1)~compare embeddings of the same size;  
(2)~treat normalization as a further hyperparameter; 
and 
(3)~use multiple classifiers during evaluation, i.e., at least logistic regression and an MLP. 
We recommend against using unsupervised cosine$+$Pearson evaluations 
but instead to use a 
learned similarity function, 
and to report MSE 
as an alternative to Pearson/Spearman correlations. If unsupervised evaluation is unavoidable, normalization is even more important. 
Finally, we think that
current transfer tasks for sentence embeddings should be complemented by more challenging ones for which bag-of-words models or random projection models cannot as easily compete.


\subsubsection*{Acknowledgments}
We thank the reviewers for their helpful feedback. 
This work has been supported by the German Research Foundation (DFG) funded research training group ``Adaptive Preparation of Information form Heterogeneous Sources'' (AIPHES, GRK 1994/1) and the DFG-funded project QA-EduInf (GU 798/18-1, RI 803/12-1). 
Some calculations of this research were conducted on the Lichtenberg high performance cluster of the TU Darmstadt. 

\bibliography{main}
\bibliographystyle{acl_natbib}

\end{document}